\documentclass[10pt,twocolumn,letterpaper]{article}

\usepackage[pagenumbers]{cvpr} 

\usepackage{graphicx}
\usepackage{amsmath}
\usepackage{amssymb}
\usepackage{booktabs}
\usepackage{pifont}
\usepackage{bm}
\usepackage{multirow}
\makeatletter

\newcommand{\Rmnum}[1]{\expandafter\@slowromancap\romannumeral #1@}
\makeatother
\usepackage{xcolor}
\usepackage{utfsym}

\usepackage[pagebackref,breaklinks,colorlinks,citecolor=cyan]{hyperref}

\usepackage[capitalize]{cleveref}
\crefname{section}{Sec.}{Secs.}
\Crefname{section}{Section}{Sections}
\Crefname{table}{Table}{Tables}
\crefname{table}{Tab.}{Tabs.}

\def\cvprPaperID{1} 
\def\confName{CVPR}
\def\confYear{2023}

\begin{document}

\title{Human 3D Avatar Modeling with Implicit Neural Representation: \\ A Brief Survey}

\author{Mingyang Sun$^{\usym{1F3F5}}$ $\quad$
        Dingkang Yang$^{\usym{1F3F5}}$ $\quad$
        Dongliang Kou$^{\usym{1F3F5}}\quad$
        Yang Jiang$^{\usym{1F3F5}}\quad$\\
        Weihua Shan$^{\usym{1F680}}\quad$ 
        Zhe Yan$^{\usym{1F680}}\quad$
        Lihua Zhang$^{\usym{1F3F5}}$\footnotemark[2] \\ 
        $^{\usym{1F3F5}}$Academy for Engineering and Technology, Fudan University$\quad$\\
        $^{\usym{1F680}}$Algorithm Innovation Lab Huawei Cloud Computing Technologies Co., Ltd Xi’an, China\\
{\tt\small 
mysun21@m.fudan.edu.cn $\{$dkyang20,lihuazhang$\}$@fudan.edu.cn
}
}
\maketitle


\renewcommand{\thefootnote}{\fnsymbol{footnote}} 
\footnotetext[2]{Corresponding Author.} 


\begin{abstract}
A human 3D avatar is one of the important elements in the metaverse, and the modeling effect directly affects people’s visual experience. However, the human body has a complex topology and diverse details, so it is often expensive, time-consuming, and laborious to build a satisfactory model. Recent studies have proposed a novel method, implicit neural representation, which is a continuous representation method and can describe objects with arbitrary topology at arbitrary resolution. Researchers have applied implicit neural representation to human 3D avatar modeling and obtained more excellent results than traditional methods. This paper comprehensively reviews the application of implicit neural representation in human body modeling. First, we introduce three implicit representations of occupancy field, SDF, and NeRF, and make a classification of the literature investigated in this paper. Then the application of implicit modeling methods in the body, hand, and head are compared and analyzed respectively. Finally, we point out the shortcomings of current work and provide available suggestions for researchers. 
\end{abstract}


\section{Introduction}
Human 3D avatars have long played an important role in industries such as video, games, and entertainment. In film-making, where actors wear costumes with sensors, motion-capture data is collected in the studio and then used to drive characters. In the game character animation, stuntmen’s movements are usually collected to produce realistic action animation. Similarly, the virtual idol is driven by the actor behind it. Recently, with the proposal of the meta-universe, the creation and generation of digital humans have attracted wide attention. Traditional methods of building 3D avatars are expensive and time-consuming. First, there needs to be a carefully arranged dedicated venue and an expensive camera array. Second, actors put on special costumes and begin to capture the movement and image of the process. Third, the production technicians need to manually correct the data generated to eliminate artifacts and fill in the missing parts. Finally, the 3D mesh will be bound to the bones, which is called skinning and directly determines the visual effect of the animation.

To reduce the manual workload, researchers try to use artificial intelligence models to recover the shape and action of people from the collected sensor data. The types of data usually include 3D point clouds\cite{bhatnagar2020combining}, RGB images, depth maps, and bone key points. The latter three are usually collected synchronously\cite{munaro20143d,ofli2013berkeley,varol2017learning}. Besides, the sensor data will be post-processed to obtain a mesh model\cite{pons2015dyna,bogo2017dynamic}. The representation of the reconstructed avatar model can be divided into three categories: voxel\cite{sharma2020voxel,varol2018bodynet}, mesh\cite{he2019data,pavlakos2018learning} and implicit field\cite{peng2021nb,saito2019pifu,mildenhall2021nerf}.
Among them, voxel and mesh are discrete representation method, which models the whole object with a large number of discrete elements. Compared with voxel, the mesh is more popular due to the convenience of representing deformation. Many studies have constructed parametric mesh models to describe various parts of the virtual avatar, including the body\cite{anguelov2005scape,loper2015smpl}, head\cite{paysan2009face,ploumpis2020towards}, and hand\cite{li2021piano,li2022nimble}.
A parametric model is a statistical model, which analyzes the common features and influencing factors of topological structure on the dataset, and establishes a general parametric equation. For example, SMPL\cite{loper2015smpl} constructs a general human body template, which includes the distribution of joints, motion constraints, shape, and posture parameters to describe personalized features. The parametric model associates personal features with the skin and bone of the 3D avatar through formulas, which greatly enhances the convenience of 3D avatar representation. However, the artificially designed parameter model has several shortcomings\cite{cheng2018parametric}: (1) It is not accurate enough, and the parameters are difficult to describe the details; (2) It is not comprehensive enough, and the model has not completely covered features such as expression and muscle; (3) It lacks photo-realistic modeling. Therefore, Mescheder \emph{et al.}\cite{mescheder2019occupancy} and Park \emph{et al.}\cite{park2019deepsdf} respectively propose two similar Implicit Neural Representation (INR) methods: occupancy filed and Signed Distance Field (SDF), whose basic idea is to treat the contour of a 3D object as a function and then uses neural networks to fit this function. Concretely, sampling a point in space, for the former, if the point is within the contour of the object, the occupancy value is 1, otherwise, it is 0, and the decision surface of the classification is the contour of the object; For the latter, the distance from the point to the contour surface is calculated. In the contour, the sign is negative, otherwise, it is positive. The plane with a distance of 0 represents the contour of the object. 
Further, inspired by SDF\cite{park2019deepsdf}, Mildenhall \emph{et al.}\cite{mildenhall2021nerf} change the contour feature to the color and the volume density and proposes the Neural Radiance Field (NeRF) which extends the INR to the rendering field and is mainly used to synthesize photo-realistic novel view images with great perspective consistency\cite{schwarz2020graf}. The methods based on INR are continuous, which can theoretically represent 3D objects with arbitrary topology and resolution, and have been widely used in 3D reconstruction\cite{peng2020convolutional}, data compression\cite{strumpler2021implicit}, image generation\cite{dupont2021generative}, and so on. In 3D avatar reconstruction, almost all of the latest excellent studies have used implicit neural networks to build more refined models. The main idea is to learn the influence of various human characteristics on blend skinning weights. There are mainly two types of contour representation: occupancy field and SDF. Researchers condition the parameters of the contour network by using features such as posture\cite{yang2021s3,mihajlovic2021leap,karunratanakul2021halo}, identity\cite{zheng2022imface,hong2022headnerf,alldieck2021imghum}, and expression\cite{zheng2022imface,alldieck2021imghum}, or model these features as implicit fields, which are decoded to influence the contour network. 

For human modeling, the current research mainly focuses on the body, hand, and head. In body-related research, inspired by the 3D reconstruction of objects, several works use the global implicit field to describe the human model. Niemeyer extends O-Net\cite{mescheder2019occupancy} and proposes O-Flow\cite{niemeyer2019occupancy} model to realize the reconstruction of an action sequence, which uses the occupancy field to describe the body contour and the velocity field to describe the deformation gradient of vertices. Yang \emph{et al.}\cite{yang2021s3} design an implicit field for the pose, skin, and shape of the avatar and makes an animated avatar. However, it will lead to an over-smooth result with a global implicit field. Many works\cite{deng2020nasa,chen2021snarf,saito2019pifu,qian2022unif} divide the body into multiple parts that are assigned an implicit field respectively and then combined. The above studies have proved that the INR is very effective in representing articulated objects, and similar methods are also applied to hand representation. LISA\cite{corona2022lisa} uses a hand skeleton model containing 16 joints, each bone had an SDF and color field, and the predicted blend skinning weights are used to weigh the output of each part to get the final result. HALO\cite{karunratanakul2021halo} also leverages the same skeleton model as LISA. LISA is based on a parameterized hand model MANO\cite{romero2017mano}, while HALO describes shapes completely based on bone information. By using bone length as a shape descriptor, HALO makes the model easier to train. In addition, some progress has been made in the implicit representation of the head, including the face and hair. HeadNeRF\cite{hong2022headnerf} extracts four kinds of latent codes of identity, expression, texture, and illumination from the image, realizing the implicit head representation considering face shape, expression, and hair. ImFace\cite{zheng2022imface} equivalently describes the 3DMMs model\cite{paysan2009face}, in which the expression and identity information is extracted and decoupled from face embedding. NeuralHDHair\cite{wu2022hdhair} introduces a hair growth model to generate the target hairstyle by building a growth orientation field and a growth occupancy field. 

Since INR is a novel concept that has been proposed in recent years, although a large number of follow-up studies have soon applied this method in various fields, there is still a lack of comprehensive review studies, especially in the representation of the human body. Chen \emph{et al.}\cite{chen2021towards} give a brief review of the progress of 3D reconstruction of the human body, but only mentioned little about INR of the human body. Cheng  \emph{et al.}\cite{cheng2018parametric} comprehensively summarize the development of parametric models for 3D human body reconstruction, but it focuses primarily on human body description and is a traditional approach rather than the latest INR. Similarly, Egger \emph{et al.}\cite{egger20203d} summarize the development of 3DMM in detail. Even though it introduces a combination of face modeling and deep learning, it still fails to cover the latest INR. Sharma \emph{et al.}\cite{sharma20223d} have done a lot of research on the use of deep learning for 3D face reconstruction, which is the latest literature, but also does not include any work related to INR at all. Bao \emph{et al.}\cite{bao2018survey} have done enough research on building hair models from images, which also do not cover INR. In a word, INR is a novel method, which has achieved better results than before in the representation of the human 3D model, but there is a lack of timely review. The main reasons are that the related studies are either not comprehensive enough\cite{chen2021towards,sharma20223d} or are outdated\cite{cheng2018parametric,sharma20223d,egger20203d,bao2018survey}. In this paper, the research on implicit neural representations for 3D avatar modeling is reviewed from three aspects: body, hand, and head. We make a summary of the latest works and analyze the powerful ability of INR. Since most of the current studies focus on the implicit representation of the body and there are few studies on the hand and head, the hand and head are summarized in the same chapter. We have also classified the literature investigated in this paper according to the method, application, and input, as detailed in Table \ref{tab:t1}.

\begin{table*}[htpb]
\setlength{\tabcolsep}{8pt}
\centering
\caption{A detailed taxonomy of the literature investigated in this paper.}
\label{tab:t1}
\resizebox{\linewidth}{!}{%
\begin{tabular}{@{}ccccccccc@{}}
\toprule$\textbf {Type}$                   & $\textbf {Publication}$                            & $\textbf {Occupancy} $ & $\textbf {SDF}$ & $\textbf {NeRF}$ & $\textbf{Part-Based}$ & $\textbf {Dynamic}$ & $\textbf {3D\,Input}$ & $\textbf {2D\,Input}$ \\ \midrule
\multirow{18}{*}{$\textbf {Body}$} & PIFu\cite{saito2019pifu}                         & $\checkmark$         &               &               & $\checkmark$         & $\checkmark$      &                    & $\checkmark$       \\
                                 & O-Flow\cite{niemeyer2019occupancy}               & $\checkmark$         &               &               &                      &                   & $\checkmark$       &                    \\
                                 & S3\cite{yang2021s3}                              & $\checkmark$         &               &               &                      &                   & $\checkmark$       & $\checkmark$       \\
                                 & NASA\cite{deng2020nasa}                          & $\checkmark$         &               &               & $\checkmark$         & $\checkmark$      & $\checkmark$       &                    \\
                                 & SNARF\cite{chen2021snarf}                        & $\checkmark$         &               &               & $\checkmark$         &                   & $\checkmark$       &                    \\
                                 & LEAP\cite{mihajlovic2021leap}                    & $\checkmark$         &               &               &                      &                   & $\checkmark$       &                    \\
                                 & PIFuHD\cite{saito2020pifuhd}                     & $\checkmark$         &               &               &                      &                   &                    & $\checkmark$       \\
                                 & Geo-PIFu\cite{he2020geopifu}                     & $\checkmark$         &               &               &                      &                   &                    & $\checkmark$       \\
                                 & SCANimate\cite{saito2021scanimate}               &                      & $\checkmark$  &               &                      &                   & $\checkmark$       &                    \\
                                 & Neural-GIF\cite{tiwari2021gif}                   &                      & $\checkmark$  &               & $\checkmark$         & $\checkmark$      & $\checkmark$       &                    \\
                                 & imGHUM\cite{alldieck2021imghum}                  &                      & $\checkmark$  &               &                      &                   & $\checkmark$       &                    \\
                                 & UNIF\cite{qian2022unif}                          &                      & $\checkmark$  &               & $\checkmark$         &                   & $\checkmark$       &                    \\
                                 & Bozic \emph{et al.}\cite{bozic2021graph}                &                      & $\checkmark$  &               & $\checkmark$         & $\checkmark$      & $\checkmark$       &                    \\
                                 & ICON\cite{xiu2022icon}                           &                      & $\checkmark$  &               &                      &                   &                    & $\checkmark$       \\
                                 & Alldieck \emph{et al.}\cite{alldieck2022photorealistic} &                      & $\checkmark$  &               &                      &                   &                    & $\checkmark$       \\
                                 & Peng \emph{et al.}\cite{peng2021anerf}                  &                      &               & $\checkmark$  &                      &                   &                    & $\checkmark$       \\
                                 & DD-NeRF\cite{yao2021ddnerf}                      &                      &               & $\checkmark$  &                      &                   &                    & $\checkmark$       \\
                                 & Neural Body\cite{peng2021nb}                     &                      &               & $\checkmark$  &                      & $\checkmark$      &                    & $\checkmark$       \\ \midrule
\multirow{3}{*}{$\textbf {Hand}$}  & HALO\cite{karunratanakul2021halo}                & $\checkmark$         &               &               &                      &                   & $\checkmark$       &                    \\
                                 & LISA\cite{corona2022lisa}                        &                      & $\checkmark$  &               & $\checkmark$         &                   & $\checkmark$       &                    \\
                                 & Grasping Field\cite{karunratanakul2020grasping}  &                      & $\checkmark$  &               &                      &                   & $\checkmark$       &                    \\ \midrule
\multirow{5}{*}{$\textbf {Head}$}  & NeuralHDHair\cite{wu2022hdhair}                  & $\checkmark$         &               &               &                      &                   &                    & $\checkmark$       \\
                                 & ImFace\cite{zheng2022imface}                     &                      & $\checkmark$  &               &                      &                   & $\checkmark$       &                    \\
                                 & Yang \emph{et al.}\cite{yang2022soft}                   &                      & $\checkmark$  &               &                      & $\checkmark$      & $\checkmark$       &                    \\
                                 & H3D-Net\cite{ramon2021h3d}                       &                      & $\checkmark$  &               &                      &                   & $\checkmark$       & $\checkmark$       \\
                                 & HeadNeRF\cite{hong2022headnerf}                  &                      &               & $\checkmark$  &                      &                   &                    & $\checkmark$       \\ \bottomrule 
\end{tabular}
}
\vspace{-5pt}
\end{table*}

The structure of this survey is as follows. In \cref{sec2}, we describe the INR methods, including the occupancy field, SDF, and NeRF. \cref{sec3} reviews the literature related to the INR of the body. The works related to hand and head are described in \cref{sec4}. In \cref{sec5}, we analyze the shortcomings of the current research and the possible development trends in the future. Finally, we summarize the survey.

\section{Implicit Neural Representation}
\label{sec2}
\begin{figure}[htbp]
  \centering
  \includegraphics[width=\linewidth]{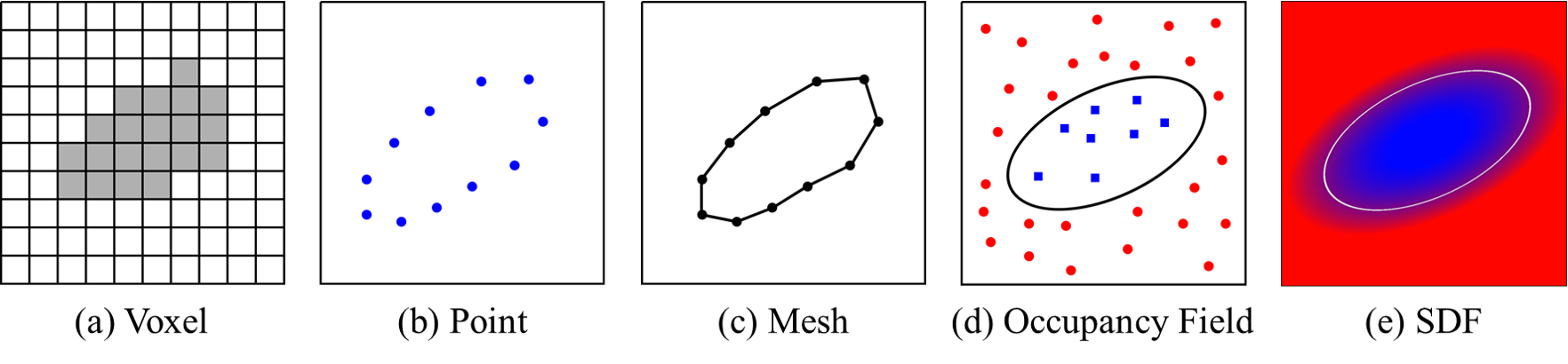}
  \caption{Examples for various representation methods. (a), (b), and (c) are discrete representations. Both (d) and (e) belong to continuous implicit neural representation.}
  \label{fig:f1}
\end{figure}

In mathematics, graph, and function are often closely related. For a simple and canonical graph, we can easily construct a function to describe its contour. But for complex shapes, such functions are often difficult to construct, especially for the presentation of details. Because of the powerful function-fitting ability of the neural network, the researchers put forward the idea of using a neural network to describe the contour of a 3D object\cite{mescheder2019occupancy,park2019deepsdf}. In Figure \ref{fig:f1}, it is obvious that the results are smoother with INR. INR is a continuous representation, which can be regarded as describing the distribution of fields in space, and it can not only describe the geometric contour but also describe any feature with fields. For example, in rendering, NeRF\cite{mildenhall2021nerf} constructs a color field and volume density field. Yang \emph{et al.}\cite{yang2022soft} construct a control signal field that is used to change facial expressions. These applications have shown that INR is more efficient than previous methods. Here, we will introduce the occupancy network, SDF, and NeRF.

\subsection{Occupancy Network}
For a point $ p\in\mathbb R^3 $ sampled in 3D space, the Occupancy Network (O-Net) maps its coordinate to an occupancy value $\left\{0,1\right\}$:
\begin{equation}
\label{eq:onet1}
    o:\mathbb R^3\to\left\{0,1\right\}.
\end{equation}

This representation requires a watertight contour, where the occupancy value of 0 means that p is outside the contour, and vice versa. Formally, the network is equivalent to a binary classification network, but we use the classification decision boundary to describe the geometric shape of the object.
With the neural network $f_\theta$, 3D point $p\in\mathbb R^3$, and ground truth observation $x\in\mathcal X$, we can further express the O-Net as:
\begin{equation}
\label{eq:onet2}
    f_\theta:\mathbb R^3\times\mathcal X\to[0,1],
\end{equation}
where $\theta$ is parameters of $ f_\theta $. The shape is implicitly defined as the 0.5-level set of the neural function:
\begin{equation}
\label{eq:onet3}
    \mathcal S = \left\{p|f_\theta(p,x)=0.5\right\}.
\end{equation}

\subsection{Signed Distance Field}
Similar to the representation of O-Net, SDF maps $p\in\mathbb R^3$ to a signed distance:
\begin{equation}
\label{eq:sdf1}
    s:\mathbb R^3\to\mathbb R.
\end{equation}
The sign represents whether the point is inside (negative) or outside (positive) the watertight surface and the amplitude represents the distance from the watertight surface. Therefore, we further express SDF as follows:
\begin{equation}
\label{eq:sdf2}
    f_\theta:\mathbb R^3\times\mathcal X\to\mathbb R,
\end{equation}
where the parameters are consistent with equation \eqref{eq:onet2}. The contour of the object is implicitly defined as the 0-level set of the neural function:
\begin{equation}
\label{eq:sdf3}
    \mathcal S = \left\{p|f_\theta(p,x)=0\right\}.
\end{equation}

\subsection{Neural Radiance Field}
NeRF is essentially consistent with SDF, which applies INR on rendering, and has achieved surprising results and received widespread attention. NeRF uses neural networks to describe the two quantities of color and volume density (opacity) required for rendering. Then a simple and differentiable numerical integration method is employed to approximate a real volume rendering step, leading to a rendering loss to optimize network parameters. The mapping process is shown as follows:
\begin{equation}
\label{eq:nerf1}
    (\sigma,c)=f_\theta(p,d),
\end{equation}
in which $d\in\mathbb S^2$ means the view of observation, $\sigma\in\mathbb {R^+}$ indicates volume density, and $c\in\mathbb R^3$ represents the RGB value. 
NeRF can rely on only sparse view input to synthesize novel views maintaining good view consistency.

\subsection{Summary}
The proposed INR greatly enhances the ability of a neural network to describe the topological structure of 3D objects. Since the observation space determines the features to be learned, we can use it to describe the specific feature by reasonably designing the observation space. In the modeling of a human 3D avatar, researchers design a vertex velocity field\cite{niemeyer2019occupancy}, blend skinning weight field\cite{peng2021anerf,mihajlovic2021leap,chen2021snarf}, hair growth field\cite{wu2022hdhair}, and so on to describe the human body in detail.

\section{INR for Body}
\label{sec3}
The body is the main part of humans. Compared with the head and hand, the research of INR for body modeling is more abundant and diverse, and different classifications can be made according to research perspectives. First, most intuitively, according to the implicit model describing the body, we can divide it into three types: occupancy-based, SDF-based, and NeRF-based which is also the organization of this section. Second, both global-based and part-based representations are suitable for modeling. Third, some studies focus on the static human body, while others on the dynamic human body. Finally, INR is able to represent either a minimally clothed body or that with clothing.

\subsection{Occupancy-Based Method}
\subsubsection{Input with 3D Information}
O-Flow\cite{niemeyer2019occupancy} is one of the earliest studies that use INR for dynamic human modeling. Though O-Net has the ability to describe the shape, for a motion sequence input, it needs to reconstruct each frame, which makes it very cumbersome in dynamic scenes. Therefore, O-Flow leverages O-Net only to generate the body shape for the first frame at one time. For the rest frames, O-Flow designs a velocity network to learn a deformation gradient field to make a description of the vertices’ trajectory on mesh, whose output will be passed into the ordinary differential equation for deformation mapping. Then the points in the material space will be mapped into the deformation space to achieve a continuous representation of the motion sequence. Although O-Flow extends O-Net to the dynamic scene representation, a fixed mesh is difficult to stably represent a large-scale change in action because it is only formed in the first frame.

\begin{figure}[htbp]
  \centering
  \includegraphics[scale=0.4]{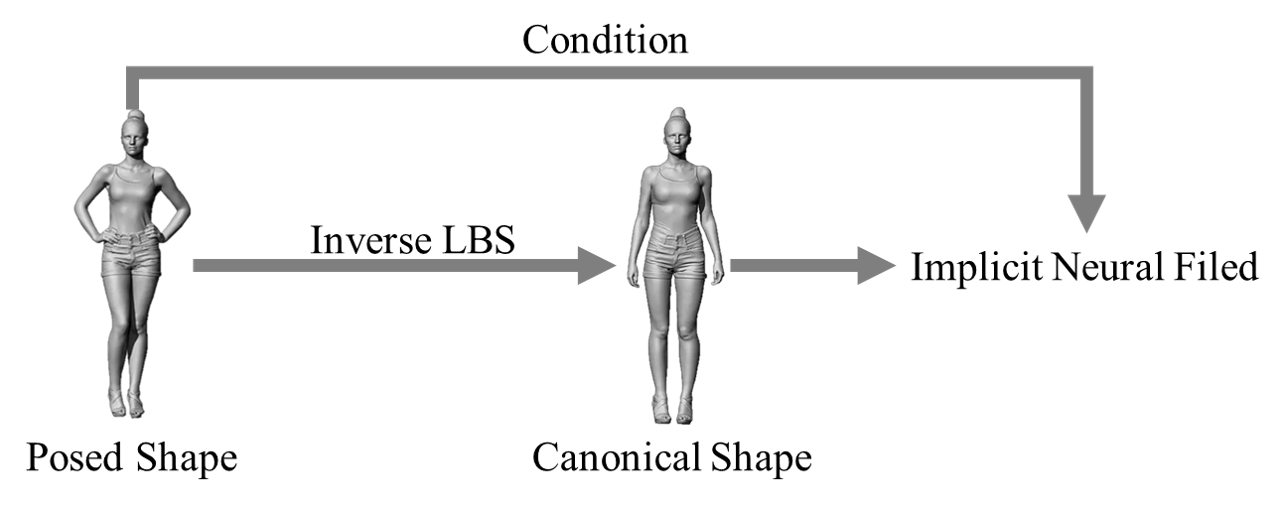}
  \caption{INR pipeline. First, the posed shape is transferred into canonical space by inverse LBS. Then, the canonical shape learned in the implicit neural field is conditioned by pose information.}
  \label{fig:f2}
  \vspace{-10pt}
\end{figure}

The idea from O-Flow inspires the following research. As shown in Figure \ref{fig:f2}, given a posed shape, researchers usually design a canonical template and map it into the canonical space using inverse Linear Blend Skinning (LBS) to learn the shape; Then, the implicit neural network is conditioned on pose information.

NASA\cite{deng2020nasa} is a part-based method, which regards the human body as an object hinged by multiple bones, and establishes an occupancy field for each bone. Different from O-Flow, NASA learns the rigid transformation of bones and local non-rigid deformation separately. First, the rigid network describes the rigid transformation of bones and relies on posture parameters to adjust the occupancy value. Second, a non-rigid deformation network arranges another operator for each bone, which is a linear subspace projection that can be learned, where the pose is projected into a low dimensional space regarded as local deformation information for shape correction. Finally, the predicted occupancy values given by each bone are sent to a maximum operator to get the result.

Similar to NASA, LEAP\cite{mihajlovic2021leap} sends the transformation matrices of all bones into the shape encoder, structure encoder, and pose encoder as global features, and then obtains their local features through the projection module of each bone. Besides, LEAP uses the transformation matrix of each bone to learn the inverse LBS, which is used to weigh the local features of all bones to condition the occupancy network. LEAP adopts the reverse skinning method that makes the learning of the canonical space pose dependent and difficult to deal with unseen poses.

SNARF\cite{chen2021snarf} avoids this problem by using the forward skinning method. Like the previous method, the first step still needs to map the posed shape into the canonical space through reverse skinning. SNARF does not directly learn the reverse skinning function but solves the inverse of the forward skinning according to the geometry cycle consistency\cite{chen2021snarf}. In addition, it also uses the iterative root-seeking method to handle the situation that the solution is not closed when the forward skin is calculated inversely.

S3\cite{yang2021s3} is a method for acquiring shape, posture, and skin features from multi-modal data. Different from the above, the input needs both 2D image and 3D voxelized radar point cloud, which are used to generate 2D feature maps and 3D feature maps respectively. There are three implicit fields, including shape field, pose field, and skin field, which are combined later to obtain an animatable 3D avatar.

\subsubsection{Input with 2D Information}

It is challenging to estimate a 3D body from a single RGB image. Saito \emph{et al.}\cite{saito2019pifu} first propose a pixel-aligned method called PIFu, which aligns 2D pixels to 3D shapes and allows the learned function to retain local details in the image. An RGB image is input and encoded by an encoder to obtain a feature map, whose pixels and the corresponding 3D vertices’ position are used to supervise the network. To improve the fidelity, PIFu leverages an RGB field to predict 3D texture. A volume occupancy field is also designed to condition the image encoder, which improves the generalization of an unseen pose. Due to the limit of hardware memory, the resolution of the input image is low, so it is difficult for PIFu to achieve high-precision reconstruction.

PIFuHD\cite{saito2020pifuhd} proposes a multi-level representation strategy. Based on PIFu, the coarse PIFu uses a normal graph of the pre-estimated front-side and back-side as additional input. Then, the fine PIFu calculates the occupancy using 3D embedding features extracted from the coarse-level network rather than absolute depth values. Though the coarse network extracts global information, the fine network does not need to know global information, so it can be trained by clipping the image to solve the memory limit problem. PIFuHD thus produces more detailed results than PIFu.

\begin{figure}[htbp]
  \centering
  \includegraphics[scale=0.285]{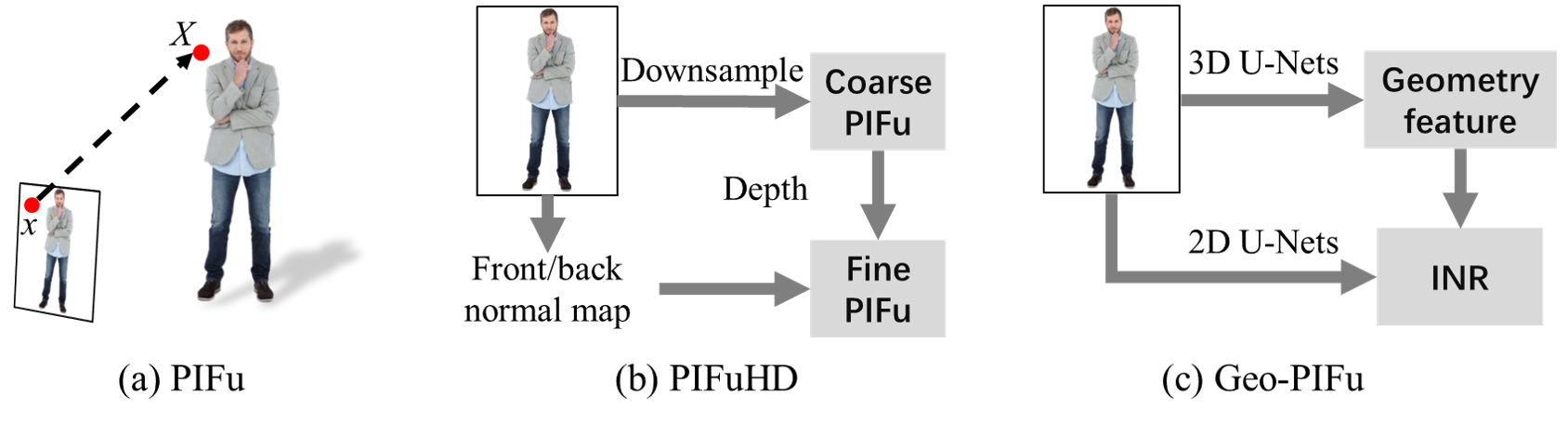}
  \caption{Comparison of pixel-aligned methods. (a) PIFu is a pixel-aligned method. (b) Based on PIFu, PIFuHD introduces a multi-level network solution to hardware memory limit. (c) Geo-PIFu combines pixel alignment with geometry alignment.}
  \label{fig:f3}
\end{figure}

PIFu uses pixel alignment to lose geometric details, while PIFuHD introduces a multi-level network of complementary details, none of which take advantage of 3D spatial information. Furthermore, Geo-PIFu\cite{he2020geopifu} improves PIFu by proposing a geometry-aligned method. They fuse 3D geometry with 2D pixels as feature vectors. Firstly, 3D U-Nets\cite{cciccek20163dunet} are used to promote pixel features to voxel features. Then, the weak perspective camera projection of a 3D point to the 2D image and a 2D encoding vector are together fed into an implicit function to describe the contour. As a consequence, the reconstruction works better for areas that are not visible in a single view. In addition, Geo-PIFu’s parameter is an order of magnitude lower than that of PIFuHD. See Figure \ref{fig:f3} for the comparison of the three methods mentioned above.

\subsection{SDF-Based Method}
\subsubsection{Input with 3D Information}

SCANimate\cite{saito2021scanimate} aims to generate a 3D avatar with adjustable posture according to given 3D scans. It is also a method of modeling the human body in the canonical space, introducing SMPL as a prior to guide the model to learn the inverse skinning weights. Similar to SNARF\cite{chen2021snarf}, SCANimate is based on the geometric cycle consistency to solve the problem that the canonical space lacks the ground truth skinning weights. The displacement of a vertex is not influenced by all joints, but only several of them. Therefore, SCANaimate introduces an attention mechanism, which effectively improves generalization.

Neural-GIF\cite{tiwari2021gif} also employs SMPL to guide the network to learn body shape. Similar to NASA\cite{deng2020nasa}, Neural-GIF divides the deformation process into two steps: rigid transformation and non-rigid deformation. For rigid transformation, they use blended skinning weights to calculate the effects of all bones. For non-rigid transformation, Neural-GIF learns a posture-dependent displacement field, while NASA adopts a projection way. Neural-GIF not only describes the contour by signed distance, but also learns the normal from the query point to the nearest surface, and further realizes the representation of the clothing details.

As a part-based representation, UNIF\cite{qian2022unif} represents the human body by 20 SDF parts. Unlike other part-based methods\cite{deng2020nasa,alldieck2021imghum}, UNIF adopts a bone-centred initialization method. Each implicit function is a sphere with a small radius centered on the bone. Then, each part continues to expand, realizing the automatic separation of the body. To avoid cracks and overlaps caused by bone rotation, they proposed a neighbour stitching algorithm, which is equivalent to a skinning method by using the rotation angle of nearby bones to affect the skinning weights of vertices.

Although UNIF realizes automatic splitting, the initialization still adopts a prior handcrafted design, which makes it difficult to generalize to other objects. To this end, Bozic \emph{et al.}\cite{bozic2021graph} propose a neural deformation graph. They use graph structure to model objects and arrange an SDF for each node to describe shapes. This is a self-supervised modeling method, and multiple loss functions are designed to make necessary constraints, including the graph structure to cover the whole body, the uniform distribution of nodes within the contour, and the sparsity of edges connections. It takes the SDF sequence as input and uses the relative displacement of graph nodes between frames to evaluate the deformation, which is then used to generate dynamic skinning weights. UNIF has strong topological adaptability and can model the dynamic interaction between the human body and deformable objects.

\begin{figure}[htbp]
  \centering
  \includegraphics[scale=0.4]{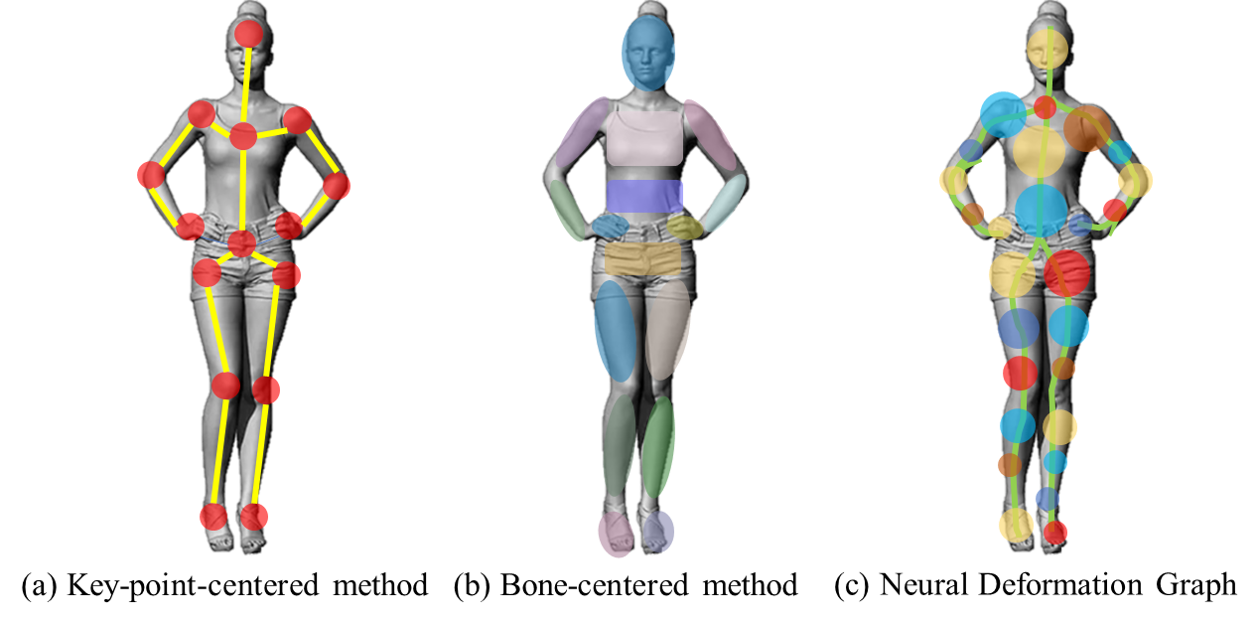}
  \caption{Schematic diagram of various part-based methods. (a) This is a key-point-centered method used by NASA\cite{deng2020nasa}, PIFu\cite{he2020geopifu}, SNARF\cite{chen2021snarf}, and Neural-GIF\cite{tiwari2021gif}. (b) Qian \emph{et al.}\cite{qian2022unif} propose a bone-centred method. (c) Tiwari \emph{et al.}\cite{tiwari2021gif} employ a self-supervised method.}
  \label{fig:f4}
\end{figure}

All of the above models only model the body, ignoring the expression and the gesture of the hands. GHUM\cite{xu2020ghum} is a parameterized model that integrates body, facial expression, and hand posture. These features are described by non-linear latent code respectively. Furthermore, imGHUM\cite{alldieck2021imghum} implements the implicit representation of GHUM and enhances expressive power. There are four SDF networks for describing the body, the left, and right hand, and facial expressions respectively.

\subsubsection{Input with 2D Information}

Alldieck \emph{et al.}\cite{alldieck2022photorealistic} believe that SDF is more informative than the occupancy field, and they remake PIFu by taking SDF to describe the shape. In addition, they estimate ambient light and surface color of reflectivity in the presence and absence of light, which allows the reconstructed model to be placed in other light scenes.

ICON\cite{xiu2022icon} takes SMPL as a prior. When inferring SMPL parameters, it employs SMPL to estimate the normal map of the front view and the back view at the same time, which are then used to infer the SDF of observable and unobservable points respectively. The normal map enhances the representation ability of clothes on the human surface, which has been proven to be effective in several studies\cite{saito2020pifuhd,he2020geopifu,tiwari2021gif}. The SMPL model inferred at the beginning must have errors, so the estimation of SMPL is iteratively improved, and the learned normal map can promote this process.

\subsection{NeRF-Based Method}

Since NeRF is mainly used to synthesize novel views, 3D reconstruction is usually a minor task or not considered. However, we aim at the 3D modeling of the human body, so the following only analyzes the content related to 3D reconstruction in NeRF-based methods.

Neural Body\cite{peng2021nb} attaches the latent code to the SMPL model as structured latent code. It is able to extract the 3D model from voxel density but with artifacts. Yao \emph{et al.}\cite{yao2021ddnerf} take SMPL priors as rough features and multi-view input as detailed features. The two features are fused to learn SDF together, which can realize 3D reconstruction and achieve a more detailed effect than Neural Body.

Inspired by the modeling pipeline implemented with an occupancy network and SDF, Peng \emph{et al.}\cite{peng2021anerf} develop NeRF for the re-animation of an avatar. They realize the canonical space with NeRF representation. Then, a neural deformation field is constructed with SMPL as a constraint, which is fed into a position encoder. The output of the position encoder is converted into the shape feature and volume density by a density model as the necessary elements of NeRF.

\label{ssec:dat}

\section{INR for Hand and Head}
\label{sec4}
Because there is less research on the modeling of hands and heads, INR will not be classified in detail in this section.
\subsection{Hand-Related Works}

HALO\cite{karunratanakul2021halo} is an implicit hand model based on an occupancy field that completely depends on bone input. Similar to the method in body modeling, the skeleton is mapped into the canonical space, whose length is regarded as the shape descriptor. For blend skinning weights, HALO keeps consistent with MANO\cite{romero2017mano}. In addition, since an object can also be represented with key points, HALO combines object code with the hand bone code. After a key point decoder, it generates the shape of grasping. This expansion is called HALO-VAE\cite{karunratanakul2021halo}. Another method\cite{karunratanakul2020grasping} describing grasps implicitly is similar to HALO-VAE, except that both the hand model and the object model are input in the form of point clouds which are passed through their occupancy network encoder respectively. The final prediction includes the distance from the query point to the hand and the object, and the decomposition of the hand. In order to avoid the penetration of the decoded object and hand, physical constraints are introduced into both methods.

LISA\cite{corona2022lisa} employs MANO for the same purpose and is a part-based method. Similar to HALO, the shape parameters in MANO are implicitly described using the skeletal posture, but LISA takes SDF to describe the shape. In addition, LISA also learned the color field to achieve a more realistic appearance.

\subsection{Head-Related Works}
NeuralHDHair\cite{wu2022hdhair} is an occupancy-based hair growth model and is able to recover a 3D hairstyle from a single image. This study assigns occupancy and orientation to points in 3D space, where occupancy describes whether the hair is growing and orientation determines the growing direction. For the occupancy field, the voxel-aligned idea is adopted, which corresponds to 2D pixels to 3D voxels. For the orientation field, the learning process is divided into two steps. First, they estimate the 3D orientation from the 2D orientation map of the original image as a rough growth direction. Second, they enrich the orientation details from the high-precision illumination map.

For face modeling, the key task is to express rich and various expressions and emotions\cite{yang2022contextual,yang2023context,yang2022emotion,yang2022disentangled,yang2022learning,yang2023target,du2021learning}. ImFace\cite{zheng2022imface} is an implicit face model, which describes the 3DMM model in a neural network equivalently. The inspiration for ImFace is similar to the implicit modeling of the human body\cite{yang2021s3}. The facial expression and identity features are extracted from the face embedding vector and decoupled. The implicit neural representation can also be used to drive facial expression. Yang \emph{et al.}\cite{yang2022soft} disperse the human face into a large number of elastic units, construct an implicit control signal field, and apply stress to each elastic unit. The change of shape is calculated by physical simulation, which provides a novel idea for the application of INR on soft-body control.

Implicit modeling based on a parametric model is limited by the original model and can not represent the whole head in detail. Recently, some achievements\cite{ramon2021h3d,hong2022headnerf} have been made in the fine modeling of the head. H3D\cite{ramon2021h3d} explores generating the entire head model from multiple views, including the hairstyle, face, and expression. First, they train an SDF with a lot of 3D scans, and then fix the parameters of the SDF network. Second, the implicit neural rendering model is trained with the current SDF as a prior. After convergence, the parameters of the SDF are unfrozen, and then the training is continued for fine-tuning. However, the reconstruction quality is affected by the pre-trained model depending on a large amount of 3D scanning data. HeadNeRF\cite{hong2022headnerf} is a head rendering model, which directly extracts four kinds of latent code from the image, including identity, expression, texture, and illumination. The first two correspond to the pixel density field, while the last two correspond to the color field. The model can interpolate latent code to generate new images. Due to differentiable rendering supported by NeRF, they can directly change the camera perspective to generate novel view images.

\section{Discussion and Suggestion}
\label{sec5}
The implicit neural field has shown great advantages in the reconstruction of human avatars. It can model any topology and a variety of features as continuous fields. Compared with traditional methods, it has greatly improved in refinement, realism, and efficiency. The occupancy field and SDF are widely used for representation. The former only outputs occupancy value, while the latter contains distance quantity. The gradient norm of the occupation field only reaches the maximum at the zero level, while that of SDF is constant. Both methods have been widely used in 3D reconstruction tasks, and they are sufficient enough to describe the shape of objects. NeRF extends the idea of SDF to the rendering and has made good progress in novel view synthesis. However, the 3D shape directly recovered from the volume density has artifacts.

Based on the investigation of the current research, we discuss several problems:
\begin{itemize}
\item Firstly, for the motion sequence reconstruction, the methods based on implicit representation only consider the spatial correlations but ignore the temporal correlations, which is also a specific motion feature.

\item Secondly, although current research has been able to model animations, the lack of physical reality has led to the performance of the avatar being heavily dependent on skinning weights and making it difficult to generalize to unseen gestures. It still remains a challenge for equipping the model with physical elasticity.

\item Thirdly, hands are of great significance for the interaction ability of an avatar. Although implicit representation has been used to describe grasping\cite{karunratanakul2020grasping,karunratanakul2021halo}, the effects are still rough, and the contact between the hand and the object can not be modeled physically.

\item Finally, for the head, hair is undoubtedly the most difficult part to simulate, even for computer graphics. Currently, the hair model is stiff, so it is valuable to use INR to establish a more refined representation of hair.
\end{itemize}

In computer graphics, impressive progress has been made in human simulation and modeling, including hair simulation\cite{li2021codimensional,fei2021mpm}, muscle simulation\cite{angles2019viper,abdrashitov2021muscle}, clothing simulation\cite{wu2022gpu,tang2018cloth}, and photorealistic hand modeling\cite{li2022nimble}. Yang \emph{et al.}\cite{yang2022soft} have taken the lead in integrating implicit neural representation and soft-body simulation and exceeded the previous work. We put forward the following suggestions:

\begin{itemize}
\item First, 
the temporal correlations of the motion sequence is further exploited by using the mechanical equations related to motion as priors or learnable parameters.


\item Second, the method based on elastic mechanics is more suitable to handle non-rigid deformation, while the skeleton is convenient to describe rigid deformation. Therefore, it is a great option to combine the elastic model and skeleton-based representation.

\item Third, in order to make the hand more expressive, the interaction between the hand and the object should be embodied in the fusion form of implicit neural fields.

\item Finally, in order to enhance the physical property, the hair model in computer graphics should serve as a prior, while the implicit shape is regarded as a physical constraint.
\end{itemize}

\section{Conclusions}
The progress of algorithms in computer science is diverse and rapid. Within a few years when INR was proposed, researchers quickly created a large number of variants according to different application scenarios. In the application of human whole-body modeling, the method based on INR has also become a trend. In this paper, we summarize the current works in time. First, we introduce the traditional modeling method and the origin of INR. Second, we expound on the relevant research related to the body, hand, and head. Finally, we propose several existing problems and corresponding opinions. In summary, INR is an excellent method that provides strong support for modeling 3D human avatars. The avatar based on implicit representation is bound to develop toward authenticity, flexibility, and personality.

\section*{Acknowledgements}
This work is supported in part by the National Key R\&D Program of China (2021ZD0113502, 2021ZD0113503), and in part by the Shanghai Municipal Science and Technology Major Project (2021SHZDZX0103).

{\small
\bibliographystyle{ieee_fullname}
\bibliography{cver}
}

\end{document}